\def\year{2022}\relax
\documentclass[letterpaper]{article} 
\usepackage{aaai22}  
\usepackage{times}  
\usepackage{helvet}  
\usepackage{courier}  
\usepackage[hyphens]{url}  
\usepackage{graphicx} 
\usepackage{natbib}  
\usepackage{caption} 
\usepackage{tikz}
\usepackage{booktabs}

\usepackage{amsmath}
\usepackage{amssymb}
\usepackage{boldline}

\usepackage{multicol}

\DeclareCaptionStyle{ruled}{labelfont=normalfont,labelsep=colon,strut=off} 
\usepackage{algorithm}
\usepackage{algorithmic}

\usepackage{newfloat}
\usepackage{listings}
\floatstyle{ruled}
\newfloat{listing}{tb}{lst}{}
\floatname{listing}{Listing}
\title{Pose Adaptive Dual Mixup for Few-Shot Single-View 3D Reconstruction}
\author{
    Ta-Ying Cheng,\textsuperscript{\rm 1,3\thanks{Denotes equal contribution.}}
    Hsuan-Ru Yang, \textsuperscript{\rm 1\textsuperscript{\thefootnote}}
    Niki Trigoni, \textsuperscript{\rm 3}
    Hwann-Tzong Chen, \textsuperscript{\rm 2}
    Tyng-Luh Liu \textsuperscript{\rm 1}
    
}
\affiliations{
    \textsuperscript{\rm 1} Institute of Information Science, Academia Sinica, Taiwan\\
    \textsuperscript{\rm 2} Department of Computer Science, National Tsing Hua University, Taiwan\\
    \textsuperscript{\rm 3} Department of Computer Science, University of Oxford, UK\\
    ta-ying.cheng@cs.ox.ac.uk, frankyang@iis.sinica.edu.tw, niki.trigoni@cs.ox.ac.uk,
    htchen@cs.nthu.edu.tw,
    liutyng@iis.sinica.edu.tw

}

\usepackage{bibentry}

\begin{document}

\maketitle

\begin{abstract}
We present a pose adaptive few-shot learning procedure and a two-stage data interpolation regularization, termed Pose Adaptive Dual Mixup (\textit{PADMix}), for single-image 3D reconstruction. While augmentations via interpolating feature-label pairs are effective in classification tasks, they fall short in shape predictions potentially due to inconsistencies between interpolated products of two images and volumes when rendering viewpoints are unknown. \emph{PADMix} targets this issue with two sets of mixup procedures performed sequentially. We first perform an \emph{input mixup} which, combined with a pose adaptive learning procedure, is helpful in learning 2D feature extraction and pose adaptive latent encoding. The stagewise training allows us to build upon the pose invariant representations to perform a follow-up \emph{latent mixup} under one-to-one correspondences between features and ground-truth volumes. \emph{PADMix} significantly outperforms previous literature on few-shot settings over the ShapeNet dataset and sets new benchmarks on the more challenging real-world Pix3D dataset.
\end{abstract}

\section{Introduction}
\begin{figure}[t!]
    \centering
    \includegraphics[width=\linewidth]{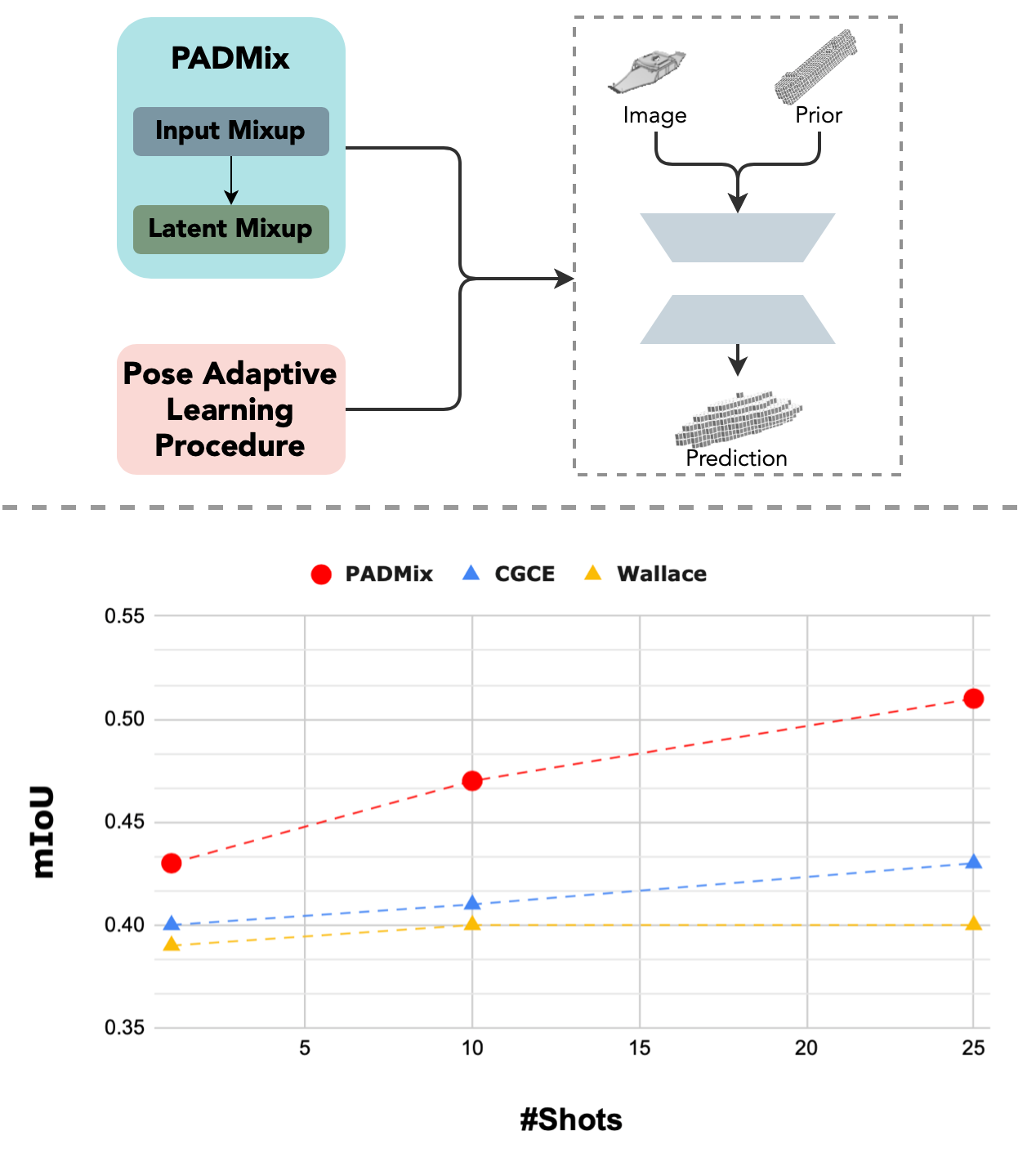}
    \caption{\textbf{Overview}. \textbf{Top:} We propose a two-stage mixup routine named \emph{PADMix} and a pose adaptive learning procedure to enhance a linear autoencoder for few-shot generalization on single-view reconstructions. \textbf{Bottom:} The mIoU scores of current methods against the number of shots. Our \emph{PADMix} performs the best at 1-shot, with a slope of improvement steeper than all previous approaches.
     }
    \label{fig:intro}
\end{figure}

\emph{Mixup}, a feature-label interpolation scheme, has been well explored and proven successful in enhancing 2D and 3D classifications \cite{mixup,pointMixup}, stabilizing generative networks, and enriching augmentations under adversarial and few-shot settings \cite{fewshotManifold}. However, literature discussing the effectiveness of interpolation regularizations in reconstructing 3D shapes from single-view images is fairly limited. We speculate that the hindrance is mostly due to the ambiguity of defining a bijective mapping between mixed inputs and outputs. Specifically, without a given pose, an interpolation between views of two objects may be inconsistent to the direct interpolation of the two objects themselves.

Aiming at transferring the benefits of \textit{mixup} on better generalizations to the reconstruction task under the challenge of pose discrepancy, we first propose a pose adaptive learning procedure on top of the effective prior-based autoencoders in few-shot reconstruction tasks \cite{fewshot3Dpriors} to promote latent representation pose invariance. With a near pose invariant encoding space, we build a two-stage data augmentation strategy, termed Pose Adaptive Dual Mixup (\emph{PADMix}) (Figure \ref{fig:intro}), to enhance the generalization of object reconstruction in novel classes with minimal training samples.

The first-stage mixup of \emph{PADMix} is performed on the input images and ground-truth volumes. We generate a training sample from an interpolation of both the 2D and 3D space of the input pairs (image and its corresponding prior), which maps to an interpolation of their two corresponding ground-truth volumes. We argue that the input mixup, while only providing a rough mapping between the mixed image and volume, is helpful in learning better feature extractions. We simultaneously impose a pose adapting loss during the input mixup training stage to minimize latent representation differences between renderings of an object from different angles.

Following the input-space mixup is a pose invariant latent mixup that can be implemented on the latent space of an autoencoder. Since the pose adaptive learning procedure enforces pose-invariance of image-prior features (i.e., latent representations of two images of the same object rendered at different angles should be similar), interpolation at this stage refines the correspondence between the mixup-generated queries and ground truths. 

Our empirical study on the popular ShapeNet dataset \cite{shapenet} shows that an image-prior encoder on par with previous work can improve significantly and achieve state-of-the-art results with the addition of \emph{PADMix}. We further explore the effects of the data-agnostic mixup procedure on scenarios with corrupted priors and no priors at all --- all of which provides consistent effectiveness of \emph{PADMix} on novel category reconstructions. Finally, we extend \textit{PADMix} to the challenging Pix3D dataset \cite{pix3d} to create a new benchmark in few-shot real-world object reconstruction.

In summary, our contributions are threefold:
\begin{itemize}

\item A pose adaptive learning procedure to promote pose invariance between latent representations of object renderings, which creates a one-to-one correspondence that could be extended for a feature-label mixup.

\item Pose Adaptive Dual Mixup (\emph{PADMix}): a data augmentation routine applicable to a 2D-3D autoencoder to enhance reconstruction results under the few-shot setting.

\item Demonstration of \emph{PADMix}’s ability to aid in model generalization and achieve state-of-the-art results on both synthesized and real-world datasets.
\end{itemize}

\section{Related Work}

\subsection{3D Reconstruction}
The process of reconstructing real-world objects from RGB images is the key to bridging 2D and 3D scene understanding. Some approaches make use of the grid nature within voxelized shape representations and build 2D-3D autoencoders based on convolutional neural networks (CNNs)~\cite{Pix2vox,Pix2vox++,corenet}. Conversely, some have focused on improving the underlying representation of 3D shapes by creating implicit functions \cite{OccupancyNetworks,reconstruction_onet}, while others emphasize on the learning of alternative 3D representations such as point clouds \cite{PSGN, lin2018learning, Mandikal2019Dense3P} and meshes \cite{Pixel2mesh,Pixel2mesh++,meshrcnn, Mask2CAD}. The idea of learning from shape priors has also been explored \cite{shapepriors,viewpriors,semantic3d,3dgan}. \citet{Yang_2021_CVPR} incorporate explicitly constructed “image-voxel” shape priors to supplement the information lost due to noisy backgrounds and heavy occlusions in the image. However, research on reconstructing 3D objects under unseen classes with limited training data remains under-developed. 

\subsection{Few-Shot Learning}
Few-shot learning is the problem of constructing models with sufficient training data from base classes and limited examples from novel classes, in the hope of learning better generalizations. Previous literature mainly focuses on 2D image tasks, mostly on classification \cite{Dhillon2020A,yu2020transmatch,afrasiyabi2020associative} and some on more complex topics such as object detection \cite{Fan_2021_CVPR,hu2021dense,FSCEv1} and segmentation \cite{PMMs2020,li2021AdaptivePL,FSSDAN}. These tasks often adopt the concept of meta-learning \cite{ren18fewshotssl,Flennerhag2020Meta-Learning,rusu2018metalearning}, where the model is trained to generalize to unseen classes in a few gradient updates.

Only few techniques focus on the predictions of 3D shapes under few-shot settings. \citet{fewshot3Dpriors} are the first to incorporate the notion of class-specific average shapes named shape priors, while \citet{fewshot3Dcompositional} propose a method to learn class-specific priors via codebooks. Our work utilizes the benefits of shape priors as secondary inputs and proposes an interpolation method for better generalization.

\subsection{Interpolation Regularization}
The notable regularization technique, referred to as \emph{mixup} \cite{mixup}, is proposed to enhance learning efficiency by generating virtual examples via interpolating an example-label pair. \citet{manifoldMixup} extend beyond this to introduce \emph{manifold mixup}, where the interpolation occurs on the hidden states instead of the inputs. These methods mainly focus on the effectiveness of regularization on 2D image tasks. \emph{PointMixup} transfers the interpolation method from grid-like pixels to 3D points and further proves that such interpolation is linear and invariant \cite{pointMixup}. Nevertheless, interpolation's effectiveness mainly shines in classification and segmentation tasks.

\emph{PADMix} derives a 2D-3D corresponding interpolation approach for shape prediction, which prevents pose variance between mixed example and ground truth to further build on the empirically-proven capability of mixup on model generalization in this new domain.

\section{Method}

\subsection{Problem Setting}

Our main objective is to learn a few-shot reconstruction model that extracts features from a 2D image $I$ containing a single object and reconstructs the corresponding 3D volume $V$. Such an approach should generalize well to novel categories of shapes with very limited training data.

In this setting of few-shot learning, training data are categorized into base categories $C_{b}$ and novel categories $C_{n}$. For every category $c \in C_{b}$, we have a set of data $\mathcal{D}_c$, where $\mathcal{D}_c = \{(I_i, V_i)\}_{i=1}^{K_c}$ and $K_c$ is the number of pairs for $c$. We also have $\mathcal{D}_{c'}= \{(I_i, V_i)\}_{i=1}^{K'}$ for every $c' \in C_{n}$. The setting is similar for $\mathcal{D}_c$ and $\mathcal{D}_{c'}$ except that $K'$ is identical across all $c'$ and  $K'\ll K_c$ for all $c \in C_{b}$.

We aim to create a training procedure and a data interpolation routine that circumvent the problem of pose discrepancy between the 2D renderings and 3D space while still encompass all the advantages of traditional interpolation regularization. Such an approach should leverage the vast quantities of $\{\mathcal{D}_c\}_{c \in C_{b}}$ and the limited samples $\{\mathcal{D}_{c'}\}_{c' \in C_{n}}$ to better generalize to the \emph{test/query} images under $C_{n}$.


\subsection{Base Network}
\begin{figure}[t]
    \centering
    \includegraphics[width=\linewidth]{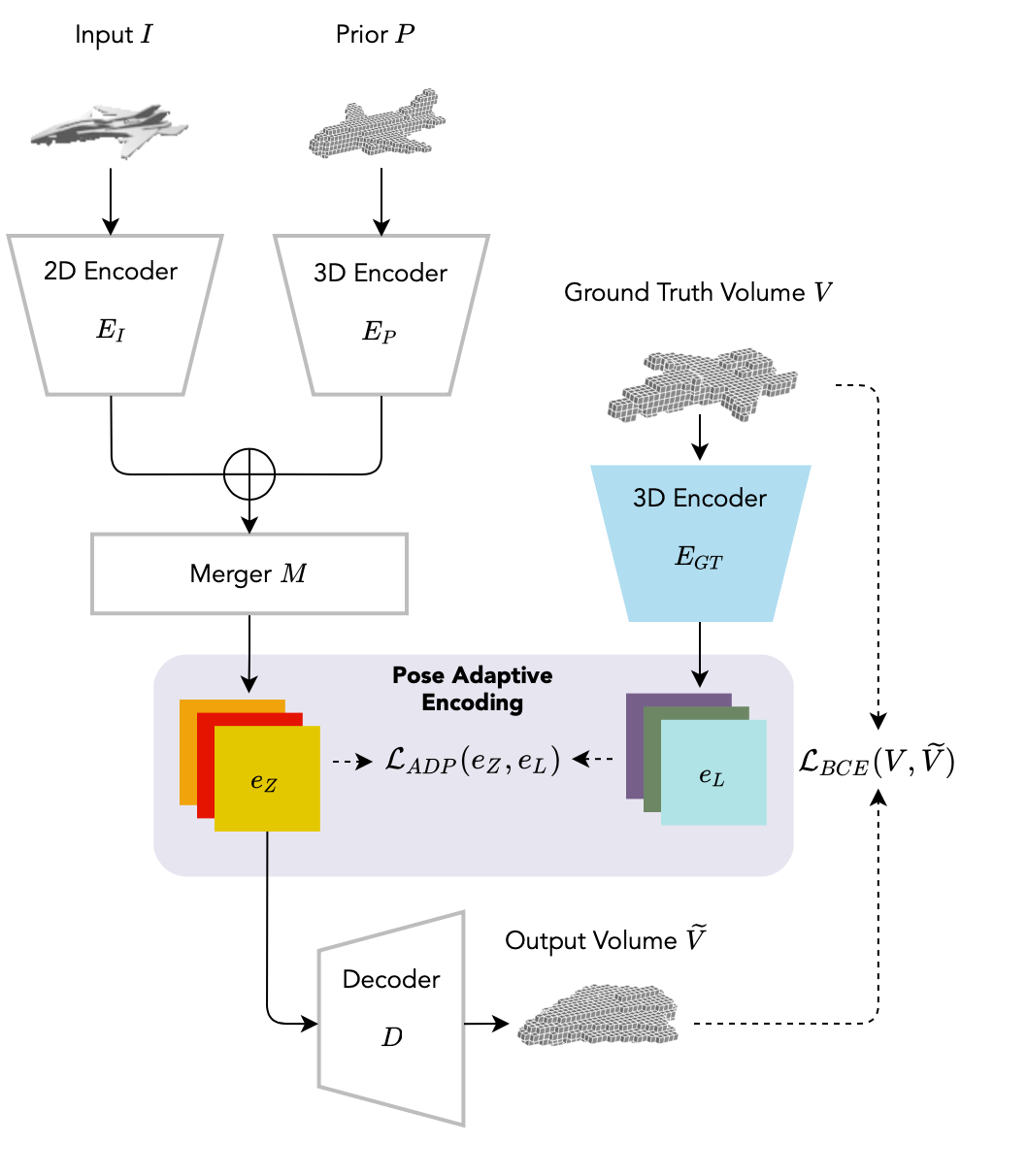}
    \caption{\textbf{Pose adaptive learning procedure}. We build upon a standard autoencoder and training pipeline for few-shot single-view reconstruction (uncolored) by introducing an additional 3D encoder to learn a pose adaptive encoding via a new pose adapting loss (colored).
     }
    \label{fig:model}
\end{figure}

We begin with a standard network architecture illustrated by the uncolored components in Figure \ref{fig:model}.
The framework comprises a 6-layer 2D image encoder $E_I$ extracted from an ImageNet-pretrained ResNet-34 \cite{imagenet,resnet}, a 4-layer 3D shape prior encoder $E_P$, a 3-layer merger network $M$, and a 4-layer decoder $D$. 

We then construct class-specific shape priors through averaging the voxel representations of 3D volumes within each class. Let $\{V_i^c\}_{i=1}^{N_c}$ be the set of $N_c$ voxel volume representations of objects under class $c$. We can construct the shape prior $P_c$ for class $c$:  
\begin{equation}
  P_c(x, y, z) = 
  \begin{cases}
  1\,, & \mbox{if } \frac{1}{N_c}\sum_{i=1}^{N_c}V_i^c(x,y,z) > t, \\
  0\,, & \text{otherwise},
  \end{cases}
  \label{eqn:prior}
\end{equation}
\noindent
where $V_i^c (x, y, z) \in \{0, 1\}$ is the value of the $i$th shape under class $c$ at voxel coordinates $(x, y, z)$ and $t$ is the binarization threshold. Previous work has empirically shown the effectiveness of naive averaging for prior generation \cite{fewshot3Dpriors}; we add thresholding so that insignificant features from the training set are omitted to avoid over-complication during our \emph{PADMix}.

Afterward, given an image $I$ and its corresponding shape prior $P$ as in (\ref{eqn:prior}), we extract image features $e_I=E_I(I)$ and shape features $e_P=E_P(P)$ through their separate encoders. We then obtain an image-prior latent representation $e_{Z}$ via the merger network $M$. Finally, $e_{Z}$  is fed into the decoder $D$ to output the final voxel-volume prediction:
\begin{equation}
    \widetilde{V} = D(e_Z) = D(M(e_I\oplus e_P)),
    \label{eqn:predV}
\end{equation}
where $\oplus$ denotes the concatenation operator.

Model learning is achieved by voxel-wise comparing our predicted volume $\widetilde{V}$ to the ground truth volume $V$. With (\ref{eqn:predV}), we let $\widetilde{V} (x, y, z) \in [0, 1]$ and $V (x, y, z) \in \{0, 1\}$ be the occupancy confidence and the ground truth label at coordinates $(x, y, z)$ respectively. We can then obtain the binary cross-entropy loss between $\widetilde{V}$ and $V$:
\begin{align}
        & \mathcal{L}_{BCE}(\widetilde{V}, V) = -\frac{1}{|V|}\sum_{(x,y,z)} [V (x, y, z)\log(\widetilde{V} (x, y, z)) \nonumber \\
 & \hspace{1.65cm} + (1 - V (x, y, z))\log(1 - \widetilde{V} (x, y, z))].
    \label{eqn:loss_BCE}
\end{align}

\subsection{Learning Pose Adaptive Encoding}

In the conventional object reconstruction settings, all images of a same object rendered at different angles should refer to the same shape prediction. This could become problematic during interpolation regularization in that the pose of a fused image may be inconsistent with its corresponding fused volume. To resolve this issue, we consider an additional shape encoder $E_{GT}$ in the training stage (exemplified by the colored components in Figure \ref{fig:model}) that maps the ground truth volume $V$ into the embedding space as $e_L$. Note that $E_{GT}$ is initialized with the encoder of a pretrained autoencoder for unsupervised volume reconstruction. The underlying goal is for $e_Z$s at all viewpoints to be as similar to $e_L$ if they refer to the same $V$. 

To achieve the above-mentioned representation alignment, we minimize the \emph{distance} between regardless of the rendering angle of $I$ by imposing a pose adapting loss comprising a triplet and a cosine similarity criterion:
    \begin{equation}
    \mathcal{L}_{ADP}(e_Z, e_L) = 
    \max(S_{zn}-S_{zp} + \mu, 0) + 1-S_{zp},
    \label{eqn:loss_ADP}
    \end{equation}
where $S_{zp}$ and $S_{zn}$ are the cosine similarities of an $e_{Z}$ generated from an image and its corresponding prior with a positive  ($e_L$ from the ground truth object) and a negative ($e_L$ from a different object in the database), and $\mu \in [0,1]$ is a margin hyperparameter. As $e_{Z}$s from all angles form positive pairs with the corresponding ground truth latent vector, we argue that $\mathcal{L}_{ADP}$ in (\ref{eqn:loss_ADP}) encourages $E_p$ to be pose adaptive via the triplet margin and cosine similarity reinforcement---outputting highly similar latent representations of images from the same object---while still preserving feature distinctiveness for images of different viewpoints. 

With a base network incorporating shape priors from previous work and a pose adaptive encoding scheme, we are ready to introduce our hierarchical \textit{PADMix} regularization.

\subsection{PADMix}
\begin{figure*}
  \includegraphics[width=\textwidth]{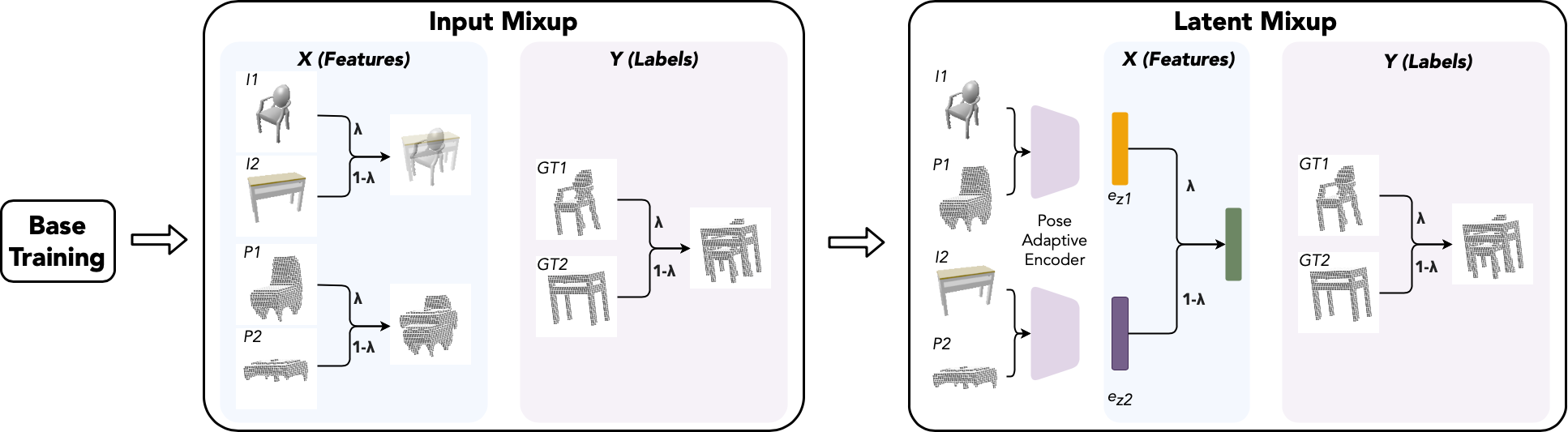}
  \caption{\textbf{Training procedure and \emph{PADMix}} augmentation routine. The overall training procedure comprises three stages: 1) Training the original base network without any data augmentation routine. 2) Training the network with input mixup to improve feature extraction and pose adaption. 3) Training the network with latent mixup. The pose invariant encoding enables better mapping between the features and the targeted volume prediction.}
  \label{fig:pipeline}
\end{figure*}

The original interpolation augmentation \emph{mixup}~\cite{mixup} is proposed to interpolate two pairs of features and their corresponding target labels $(X_1, Y_1)$ and $(X_2, Y_2)$ with a mixup ratio $\lambda \sim Beta(\alpha, \alpha)$ for $\alpha \in [0, 1]$:
    \begin{align}
        X_{mix}(\lambda) &= (1 - \lambda)X_1 + \lambda \, X_2, \label{eqn:X_mix}\\
        Y_{mix}(\lambda) &= (1 - \lambda)Y_1 + \lambda \, Y_2. \label{eqn:Y_mix}
    \end{align}
 
We are thus motivated by establishing \emph{PADMixup} via a two-stage routine that \emph{mixup} is  hierarchically carried out, first in the input space and then in the latent space, to maximize the results of reconstruction. 

The intuition behind \emph{PADMixup} is simple: 
While applying \emph{mixup} to create virtual examples via cross-class interpolations has been shown to be useful for dealing with classification and segmentation problems, it is reasonable to expect that its effective generalization could bring improvements in tackling more challenging learning scenarios such as novel shape inference from 2D images under few-shot setting.

\subsubsection{Input Mixup}
        From (\ref{eqn:X_mix}) and (\ref{eqn:Y_mix}), input mixup is achieved by naive interpolations of inputs $X = (I, P)$ and of their target volumes $Y = V$. When $X_1$ and $X_2$ are of different classes, mixup would result in a new prior. Such interpolations can be viewed as yielding a virtual class of objects with the newly interpolated prior and image, which can be thought of as a virtual example of the class.
        
        This approach, while straightforward, may cause inconsistencies between $X_{mix}=(I_{mix}, P_{mix})$ and $Y_{mix}$. Theoretically, a perfect $I_{mix}$ should be a particular view of the interpolated ground truth volume $Y_{mix}$. However, as the poses of the original renderings are usually not given, $I_{mix}$ may be inconsistent with the fused $Y_{mix}$ and $P_{mix}$ (e.g., the mixed image may have the chair facing left and table right, but the fused volume have both of them facing front).
        
        Nevertheless, we hypothesize that input mixup still has its merits on account of two main reasons. First, interpolation has been proven successful in enhancing feature extractions, which is helpful for finer reconstructions. Second, shape priors are generated from ground truth target volumes, meaning that the interpolated outcome of the two would remain consistent and thus contains implicit information about the pose of ground truth. Extensive studies to address these claims are presented in the following section.

            
            

    \subsubsection{Latent Mixup}
    With a well-trained pose adaptive encoder, we then propose a latent mixup where the input $(I,P)$s are now replaced by $e_Z$s. Since the pose adaptive encoding minimizes the cosine distances between image-views and ground truth volume representations, latent vectors $e_Z$s are implicitly distilled for being pose invariant. That is, the images from different viewpoints of a same object should be highly similar when transformed to the latent representations. This design creates a one-to-one mapping between the features and outputs, making the mixup augmentation more straightforward for networks to learn.
    
    \subsection{PADMix Training}
    A hierarchical training procedure for \emph{PADMix} (Figure \ref{fig:pipeline}) is described as follows. We first train the base network via the loss $\mathcal{L}$, accounting for both (\ref{eqn:loss_BCE}) and (\ref{eqn:loss_ADP}):
    \begin{equation}
        \mathcal{L} = w_{BCE} \cdot \mathcal{L}_{BCE} + w_{ADP} \cdot \mathcal{L}_{ADP},
    \end{equation}
    where $w_{BCE}$ and $w_{ADP}$ are hyperparameters. A complete input mixup routine is then added to the training procedure, which has been empirically shown to outperform beginning with input mixup from scratch.

    Afterward, we continue training with latent mixup. The stagewise training ensures that our latent mixup is built upon a well-trained pose adaptive encoder and that our latent representations are near pose invariant. As the interpolation takes place after the pose adaptive encoding, the definition of a positive pair is ambiguous and so we omitted the triplet criterion in $\mathcal{L}_{ADP}$ during this stage of training.

\section{Experiments}

We extensively study the generalization results of \emph{PADMix} on the ShapeNet dataset~\cite{shapenet}, following the identical settings as previous work in the 80-20 split of base classes \{airplanes, cars, chairs, displays, phone, speakers, tables\} and novel classes \{cabinet, sofa, bench, watercraft, rifle, lamp\}. All procedures are trained using eight Nvidia Tesla V100s for 100 epochs with a batch size of 32. In terms of hyperparameters, $\mu$ is set to 0.1, $\alpha$ to 0.2, and $w_{BCE}$ and $w_{ADP}$ to 10 and 0.5. The learning rates of the entire base network and the additional shape encoder $E_{GT}$ are set to 1e-3 and 1e-4, respectively.  Our main comparisons are with \citet{fewshot3Dpriors} who first introduced priors and CGCE by  \citet{fewshot3Dcompositional} that incorporates codebooks to learn better priors. 

We also extend \emph{PADMix} to the more challenging Pix3D dataset \cite{pix3d}. The dataset is an extension from the IKEA furniture dataset~\cite{ikea}, with 395 3D shapes mapping to over 10K real-world images correspondingly. Our results set a new benchmark for in-the-wild few-shot single-view reconstructions.  

Additional training details can be found in the supplementary materials.

\subsection{Results and Ablation Study}
\subsubsection{Few-shot Generalization on ShapeNet}

 \begin{table}[t]
     \resizebox{0.99\columnwidth}{!}{%
    \begin{tabular}{lcccc}
    \toprule
    \textbf{Category} & \multicolumn{1}{l}{WithoutMix} & \multicolumn{1}{l}{InputMix} & \multicolumn{1}{l}{LatMix} & \multicolumn{1}{l}{\emph{PADMix}} \\
    \cmidrule(lr){2-5}
    Cabinet    & 0.63	& 0.63	 & 0.64	& \textbf{0.67}	 \\
    Sofa       & 0.51	& 0.52	 & 0.51	& \textbf{0.54}                \\
    Bench      & 0.30	& 0.32	 & 0.33	& \textbf{0.37}                \\
    Watercraft & 0.40	& 0.40	 & 0.40	& \textbf{0.41}                \\

    Lamp       & 0.27	& 0.27	 & 0.28	& \textbf{0.29}                \\
    Rifle      & 0.34	& 0.34	 & \textbf{0.35}    & 0.31	            \\
    \cmidrule(lr){2-5}
    Average    & 0.41	& 0.41	 & 0.42	& \textbf{0.43}                \\   
    \bottomrule
    \end{tabular}
    }
    \caption{\textbf{Few-shot learning IoU results on novel ShapeNet classes}. We sequentially perform the procedures of \emph{PADMix} to see the incremental improvements of the pipeline. Bold texts denote best results.}
    \label{tab:table-1} 
    \end{table}

\begin{table*}[t]
\centering
\setlength\tabcolsep{4pt} 
\begin{tabular}{lccccccccc} 

\toprule
& \multicolumn{3}{c}{\textbf{1-Shot}} & \multicolumn{3}{c}{\textbf{10-Shot}} & \multicolumn{3}{c}{\textbf{25-Shot} }\\
 \cmidrule(lr){2-4} \cmidrule(lr){5-7} \cmidrule(lr){8-10} 
   \textbf{Category} & Wallace & CGCE & \emph{PADMix} &  Wallace & CGCE & \emph{PADMix} & Wallace & CGCE & \emph{PADMix}\\
\cmidrule(lr){2-4} \cmidrule(lr){5-7} \cmidrule(lr){8-10} 

Cabinet & 0.69 & \textbf{0.71} & 0.67 & 0.69 & \textbf{0.71} & 0.66 & 0.69 & \textbf{0.71} & 0.68\\
Sofa & \textbf{0.54} & \textbf{0.54} & \textbf{0.54} & 0.54 & 0.54 & \textbf{0.57} & 0.54 & 0.55 & \textbf{0.59}\\
Bench & \textbf{0.37} & \textbf{0.37} & \textbf{0.37} & 0.36 & 0.37 & \textbf{0.41} & 0.36 & 0.38 & \textbf{0.42}\\
Watercraft & 0.33 & 0.39 & \textbf{0.41} & 0.36 & 0.41 & \textbf{0.46} & 0.37 & 0.43 & \textbf{0.52}\\
Lamp & 0.20 & 0.20 & \textbf{0.29} & 0.19 & 0.20 & \textbf{0.31} & 0.19 & 0.20 & \textbf{0.32}\\
Rifle & 0.21 & 0.23 & \textbf{0.31} & 0.24 & 0.23 & \textbf{0.39} & 0.26 & 0.28 & \textbf{0.50}\\
\cmidrule(lr){2-4} \cmidrule(lr){5-7} \cmidrule(lr){8-10} 
Average & 0.39 & 0.41 & \textbf{0.43} & 0.40 & 0.41 & \textbf{0.47} & 0.40 & 0.43 & \textbf{0.51}\\

\bottomrule
\end{tabular}
\caption{\textbf{Class-wise IoU comparison of \emph{PADMix} to state-of-the-art few-shot learning approaches} on ShapeNet. Bold texts denote best results. \emph{PADMix} has shown to be the most effective in 1, 10, and 25-shot settings, with a widening difference margin as the number of shots increases.}
\label{tab:main}
\end{table*}

\begin{figure}[t]
    \centering
  \includegraphics[width=\linewidth]{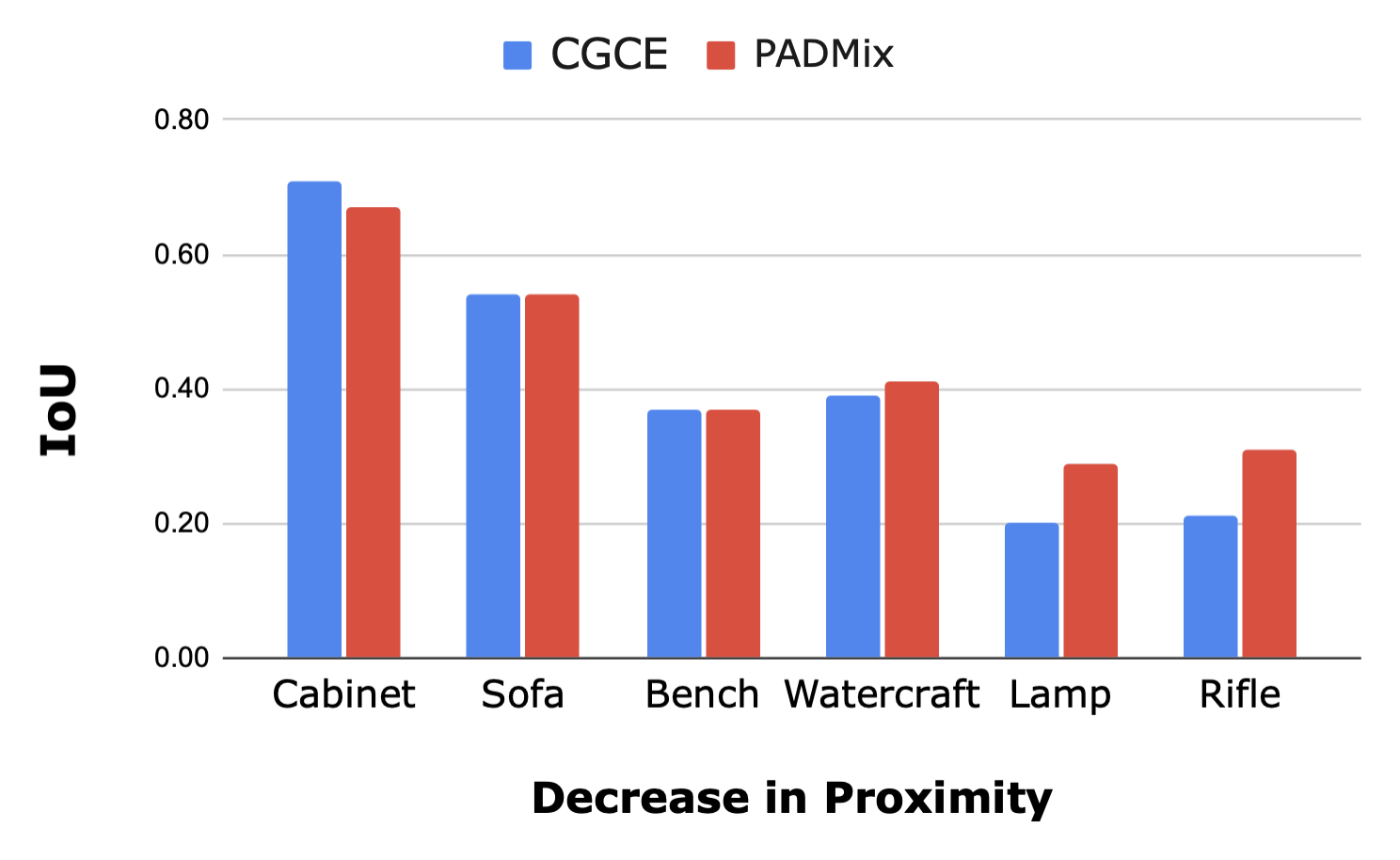}
  \caption{\textbf{1-shot class-wise IoU for decreasing proximity}. The improvement margin enlarges as proximity of the novel class against base class decreases.}
  \label{fig:proximity}
\end{figure}

We first evaluate the sequential improvements of \emph{PADMix} on our linear autoencoder (Table \ref{tab:table-1}) based on the class-wise Intersection-over-Union (IoU) metric under the 1-shot setting. While the results of input mixup and latent mixup performed separately mildly improve from the base network with pose adaptive training, the entire \emph{PADMix} achieves the best results in five out of six categories and on the overall average.

We report the best \emph{PADMix} IoU scores in comparison with results directly quoted from previous work~\cite{fewshot3Dcompositional,fewshot3Dpriors} in Table \ref{tab:main}. In all three of their given settings (1, 10, 25), \emph{PADMix} achieves higher average IoUs and outperforms in five of the six novel classes, with widening gap as the number of shots increases. In fact, our 1-shot results are comparable with the previous state-of-the-art IoUs trained under the 25-shot setting.

One observation, however, is that \emph{PADMix} tends to perform better in shapes with lower \textit{proximity} to the shapes in $C_b$ (e.g., cabinets). We further analyze this by plotting our 1-shot IoU results against proximity of each novel class against base classes in Figure \ref{fig:proximity}, where the proximity of each novel class $c$ follows the definition of \cite{fewshot3Dcompositional} and is computed as:
\begin{equation}
    Prox_c  = \frac{1}{N_c}\sum_{i=1}^{N_c} \max_{j \in BaseShapes}(IOU(V_i, V_j)).
\end{equation}
A trend of increasing margin can be observed as proximity of the novel class decreases. We hypothesize this to be the result of our $\mathcal{L}_{ADP}$ design. 

Intuitively, $\mathcal{L}_{ADP}$ encourages a difference greater than $\mu$ in between the representation of every object. When objects are physically similar and refer to the same class priors, the encoder may accommodate feature details of images to create the difference margin. Consequently, the model emphasizes the detailed feature differences rather than the global similarity of objects. Novel class objects exhibiting distinctive features are hence benefited more from our approach than objects with higher proximity to base classes. 

We argue that this setting is actually preferable, as real-world object classes tend to be diverse and highly dissimilar to one another.


\subsubsection{Qualitative Analysis}
\begin{figure}[t]
    \centering
  \includegraphics[width=\linewidth]{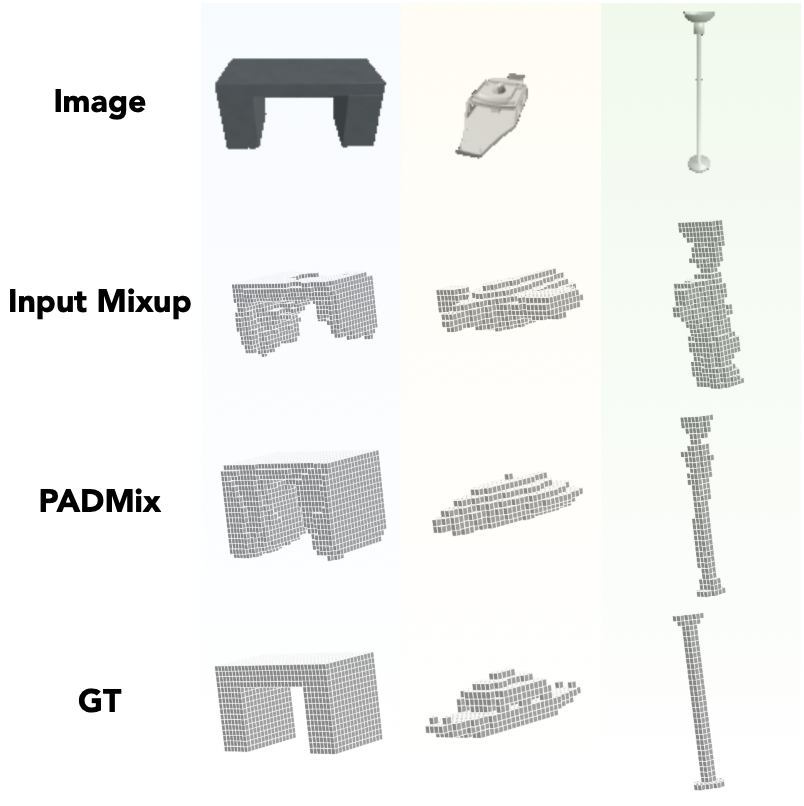}
  \caption{\textbf{Visualizations under 1-shot setting}. \textbf{Left column}: simple shape and angle. \emph{PADMix} refines the shape into a better reconstructed volume. \textbf{Middle column}: simple shape but sub-optimal angle. \emph{PADMix} produces a well-reconstructed result while Input Mixup implicitly recognizes and reconstructs the image as a plane (base class). \textbf{Right column}: difficult shape. Input Mixup fails to extract the shape features which \emph{PADMix} is able to accomplish.}
  \label{fig:vis}
\end{figure}

We juxtapose the generated outputs of input mixup and \emph{PADMix} in Figure \ref{fig:vis}. Our visualizations suggest that at times when the angle of the object or the object itself makes the reconstruction task inherently difficult, \emph{PADMix} generalizes significantly better than just the input mixup. For cases that are simpler, \emph{PADMix} also shows more refined results in terms of the overall shapes.

\subsubsection{Pose Adaptiveness}
\label{sec:poseadp}
  \begin{table}[t]
  \centering
  \resizebox{0.99\columnwidth}{!}{%
    \setlength\tabcolsep{2pt} 
    \begin{tabular}{lcccccc}
    \toprule
    & \multicolumn{2}{c}{\textbf{Base (No $\mathcal{L}_{ADP}$)}} & \multicolumn{2}{c}{\textbf{Base (With $\mathcal{L}_{ADP}$)}}& \multicolumn{2}{c}{\textbf{InputMix}}\\
    \cmidrule(lr){2-3} \cmidrule(lr){4-5} \cmidrule(lr){6-7}
     \textbf{Category}  & SameObj & DiffObj & SameObj & DiffObj & SameObj & DiffObj \\
    \cmidrule(lr){2-3} \cmidrule(lr){4-5} \cmidrule(lr){6-7}   
    Cabinet    &     0.86    &  0.62       &    0.97      &     0.84 & 0.97 & 0.85     \\
    Sofa       &    0.82     &  0.72       &    0.92      &     0.86 & 0.93 & 0.88    \\
    Bench      &    0.81     &  0.72       &    0.94      &     0.88 & 0.95 & 0.89     \\
    Watercraft &    0.84     &  0.75       &    0.96      &     0.90 & 0.96 & 0.91     \\
    Lamp       &    0.91     &  0.69       &    0.97      &     0.79 & 0.97 & 0.81     \\
    Rifle      &    0.83     &  0.80       &    0.95      &     0.93 & 0.96 & 0.95     \\
    \bottomrule
    \end{tabular}
    }
    \caption{\textbf{Cosine similarities of latent representations}. We compute the average cosine similarities between latent vectors of two images from identical and different objects.}
    \label{tab:cossim}
    \end{table}

We explicitly analyze the effectiveness of $\mathcal{L}_{ADP}$ and input mixup in promoting pose invariance and feature distinctiveness for the latent representation; we train three networks: two base networks with and without $\mathcal{L}_{ADP}$ , and one with an additional input mixup. We then compute the intra-class cosine similarities under two settings:
\begin{itemize}
\item Same Object (SameObj): where two images are rendered from the same object but at different viewpoints.
\item Different Objects (DiffObj): where two images are rendered from different objects but within the same class.
\end{itemize}
We perform the experiment within classes, which is a more challenging setting as the similarities between object volumes are higher.

Based on our results in Table \ref{tab:cossim}, $\mathcal{L}_{ADP}$ and input mixup enhance the cosine similarities under the SameObj setting across all categories, implying a better learned pose adaptive encoding. On the other hand, the margin between the two settings remained, and therefore we claim that the distinctive features for distinguishing objects are still preserved despite the enhancement in pose invariance.

\subsubsection{PADMix With No/Corrupted Priors}
    \begin{table}[t!]
     \resizebox{0.99\columnwidth}{!}{%
    \begin{tabular}{lcccc}
    \toprule
    & \multicolumn{2}{c}{\textbf{No Priors}} & \multicolumn{2}{c}{\textbf{Corrupted Priors}}\\
    \cmidrule(lr){2-3} \cmidrule(lr){4-5}
     \textbf{Category}  & WithoutMix & \emph{PADMix} & WithoutMix & \emph{PADMix} \\
    \cmidrule(lr){2-3} \cmidrule(lr){4-5}       
    Cabinet    &    0.68    &    0.68     &    0.67      &      0.67    \\
    Sofa       &    0.53    &    0.55     &    0.51      &      0.51    \\
    Bench      &    0.37    &    0.39     &    0.35      &      0.38    \\
    Watercraft &    0.36    &    0.37     &    0.36      &      0.38    \\
    Lamp       &    0.27    &    0.27     &    0.26      &      0.26    \\
    Rifle      &    0.18    &    0.19     &    0.18      &      0.17    \\
     \cmidrule(lr){2-3} \cmidrule(lr){4-5}     
    Average    &    0.40    &    0.41     &    0.39      &      0.40   \\
    \bottomrule
    \end{tabular}
    }
    \caption{\textbf{Reconstruction results with no/corrupted priors}. We adjust our base network into not feeding in any class-specific priors or priors from a different class to see the effect of \emph{PADMix} under more challenging scenarios.}
    \label{tab:prior}
    \end{table}

Following previous literature, all our experiments have the presumed knowledge of the input image class so that a correct prior is chosen. Thus, we proceed to explore the effects of \emph{PADMix} in the circumstances where the ground-truth categorical information of the input image is absent, by simulating situations with inputs consisting of no priors and corrupted priors. In the no-prior setting, we remove the 3D encoder and add an adaptive pooling on the output of the 2D encoder to readjust the feature size to fit into the merger. In the corrupted prior setting, we deliberately select a wrong prior (i.e., prior from other classes). We train all networks under the 1-shot setting with identical hyperparameters.

As indicated in Table \ref{tab:prior}, \emph{PADMix} achieves higher IoU results in all novel classes under the no-prior setting. This suggests that the interpolation regularization at both the input and latent stages could help out with extracting important features that may be well generalized to unseen objects.

The results on corrupted priors are coherent with the aforementioned findings. Even under the situation where priors are fundamentally flawed, \emph{PADMix}'s ability to extract image features has aided in better reconstruction results.

\subsubsection{Variations of Beta Distributions}
    \begin{table}[t!]
    \centering
    \begin{tabular}{lcc}
    \toprule
     & \textbf{InputMix}& \textbf{LatentMix}\\
     \cmidrule(lr){2-3}
            $\alpha =0.2$    & \textbf{0.41} & 0.41 \\
            $\alpha =0.4$    & \textbf{0.41} & \textbf{0.43} \\
            $\alpha =1$    & 0.34 & 0.41 \\
    \bottomrule
    \end{tabular}
    \caption{\textbf{Class-wise average IoUs on varying $\beta(\alpha, \alpha)$s}. Bold texts denote best results for each mixup stage.}
    \label{tab:beta}
    \end{table}

The type of distribution to use for interpolation weights is highly important. We examine the results of input mixup and latent mixup using different values of $\alpha$ for $\beta(\alpha, \alpha)$. Since the mixup procedure is sequential, we carry out testings on input mixup first, and then use the best results to test on latent mixup.

It could be concluded from Table~\ref{tab:beta} that out of the three more popular settings, input mixup achieves the best IoU under the settings $\alpha=0.2, 0.4$, while latent mixup achieves the best IoU results with $\alpha=0.4$. It is worth noting that the interpolation results in both cases fall when $\alpha=1$ (a uniform distribution). This is reasonable as equal weights of two inputs create more difficult and unrealistic examples, which should not have an equal chance of existence with examples having one dominant input.

\subsubsection{Pix3D Benchmark}
\begin{table}[t!]
\footnotesize
\centering
\begin{tabular}{l ccc}
\toprule
\midrule
\textbf{Category}  & Wallace & WithoutMix & \emph{PADMix}  \\
\cmidrule(lr){2-4}
Sofa &  0.26 & \textbf{0.39}    & \textbf{0.39} \\
Desk &  0.06 & 0.05    & \textbf{0.11}\\
 \midrule
 \midrule
\textbf{Category}  & Wallace & WithoutMix & \emph{PADMix}  \\
\cmidrule(lr){2-4}
Sofa &  0.34 & 0.38 &  \textbf{0.40}\\
Desk &   0.06 &  0.08& \textbf{0.12}\\
Bookcase & 0.05 &0.03& \textbf{0.06}\\
Misc &  \textbf{0.10} &\textbf{0.10}& 0.09\\
\bottomrule
\end{tabular}
    \caption{\textbf{Class-wise IoU results on Pix3D.} We provide benchmarks of two data splits. Benchmark 1 focuses on the model's generalization to similar training classes. Benchmark 2 provides a more comprehensive overview of the model's few-shot results. Bold texts denote best results.}
    \label{tab:pix3d}
\end{table}

Few-shot settings aim to mimic a realistic scenario of object reconstruction. With this in mind, we extend our approach to the more challenging dataset---Pix3D \cite{pix3d}---that uses in-the-wild instead of synthetic images. The volume resolution of Pix3D ($128^3$) is also much higher than that of ShapeNet ($32^3$), making the task considerably more difficult by nature.

As Pix3D is substantially smaller than ShapeNet (only around 10K images and 400 models), and with some classes only containing a dozen of models, we only provide benchmarks under the 1-shot setting. We extract all training and testing data from the standard $S_1$ split described in the Mesh-RCNN paper \cite{meshrcnn}, which contains 7539 train images and 2530 test images. Since Pix3D is loosely annotated (i.e., one image may contain more than one object but only one is labeled), we use the ground-truth bounding boxes to crop out all images. We create two benchmarks to target different aspects of reconstruction: 
The first benchmark includes only four of the nine classes: \{chair, table\} for base and \{sofa, desk\} for novel. This serves as a simpler baseline, with chairs being similar to sofas and tables to desks. This benchmark tests the ability of one's model in generalizing to new classes with high proximity to the training set.
The second benchmark serves as a more general baseline comprising all the nine classes, where we set five of the classes \{wardrobe, bed, tool, chair, table\} to base and the other four \{bookcase, desk, sofa, miscellaneous\} to novel. The results on this benchmark should be a more comprehensive overview toward one's reconstruction model.

The reported results are obtained by following the same hyperparameters for training, with one amendment made on the reconstruction loss: we observe that in reconstruction it is empirically better to use a balanced focal loss instead of the BCE loss. This could be due to a class imbalance between occupied and empty volumes (Pix3D objects are much more irregularly shaped with more empty spaces), akin to the foreground-background class imbalance that the focal loss is originally designed to solve.
Table~\ref{tab:pix3d} shows improvements for both settings in almost all classes. The extra base-categories in Benchmark 2 also allow better generalization in the overlapping categories (i.e., sofa and desk). Nevertheless, the results in Bookcase and Miscellaneous suggest that there are still plenty of rooms worth exploring.

\section{Conclusion}
This paper explores the extent of interpolation regularization in few-shot shape prediction problems. We propose a few-shot learning procedure followed by an augmentation routine named \emph{PADMix} that involves two mixup schemes: an input mixup and a pose invariant latent mixup. The former, combined with a pose triplet-cosine-based loss, strengthens the pose-adaptiveness of the encoders while maintaining feature discrepancies between different objects. The latter makes use of such pose invariance to perform a one-to-one interpolation regime between the features and labels (i.e., targeted volume). Our state-of-the-art few-shot results on the synthesized ShapeNet and real-world Pix3D datasets justify that interpolation augmentations can be well-adopted into the domain of shape predictions.

\section{Acknowledgments}
This work was supported in part by the MOST grants 110-2634-F-001-009, 110-2634-F-007-027 and 110-2221-E-001-017 of Taiwan. It was also partly supported by the ACE-OPS grant EP/S030832/1. We are grateful to National Center for High-performance Computing for providing computational resources and facilities.



\bibliography{aaai22.bib}

\end{document}